\documentclass[runningheads]{llncs}

\usepackage{eccv}

\usepackage{eccvabbrv}

\usepackage{graphicx}
\usepackage{booktabs}
\usepackage{graphicx}
\usepackage{amsmath}
\usepackage{amssymb}

\usepackage{booktabs}
\usepackage{colortbl} 
\usepackage{xcolor}
\usepackage{bigstrut}
\usepackage{multirow}
\usepackage{rotating} 
\usepackage{diagbox}
\usepackage{color}
\usepackage{marvosym}

\usepackage{wrapfig}

\usepackage[colorinlistoftodos]{todonotes}

\usepackage[accsupp]{axessibility}  

\usepackage[pagebackref,breaklinks,colorlinks,citecolor=eccvblue]{hyperref}

\usepackage{orcidlink}
\usepackage{ifsym}

\begin{document}

\title{Boosting Medical Image-based Cancer Detection via Text-guided Supervision from Reports} 

\titlerunning{Boosting Medical Image-based Cancer Detection}

\author{Guangyu Guo\inst{1,2,3}\thanks{Work was done during an internship at Alibaba DAMO Academy.  \\ \noindent $^{\textrm{\Letter}}$ Correspondence to Jiawen Yao and Dingwen Zhang.},
Jiawen Yao\inst{1,3,}$^{\textrm{\Letter}}$ \and
Yingda Xia\inst{1} \and Tony C. W. Mok\inst{1,3} \and Zhilin Zheng\inst{1,3} \and Junwei Han\inst{2} \and Le Lu\inst{1} \and Dingwen Zhang\inst{2,}$^{\textrm{\Letter}}$ \and Jian Zhou\inst{4,5} \and Ling Zhang\inst{1}}

\authorrunning{G. Guo et al.}

\institute{DAMO Academy, Alibaba Group \and
 Hefei Comprehensive National Science Center, Hefei, China \and Hupan Lab, 310023, Hangzhou, China \and
 Sun Yat-sen University Cancer Center, Guangzhou, China \and South China Hospital, Medical School, Shenzhen University, Shenzhen, China
 } 

\maketitle

\begin{abstract}

The absence of adequately sufficient expert-level tumor annotations hinders the effectiveness of supervised learning based opportunistic cancer screening on medical imaging. Clinical reports (that are rich in descriptive textual details) can offer a ``free lunch'' supervision information and provide tumor location as a type of weak label to cope with screening tasks, thus saving human labeling workloads, if properly leveraged. However, predicting cancer only using such weak labels can be very changeling since tumors are usually presented in small anatomical regions compared to the whole 3D medical scans. Weakly semi-supervised learning (WSSL) utilizes a limited set of voxel-level tumor annotations and incorporates alongside a substantial number of medical images that have only off-the-shelf clinical reports, which may strike a good balance between minimizing expert annotation workload and optimizing screening efficacy. In this paper, we propose a novel text-guided learning method to achieve highly accurate cancer detection results. Through integrating diagnostic and tumor location text prompts into the text encoder of a vision-language model (VLM), optimization of weakly supervised learning can be effectively performed in the latent space of VLM, thereby enhancing the stability of training. Our approach can leverage clinical knowledge by large-scale pre-trained VLM to enhance generalization ability, and produce reliable pseudo tumor masks to improve cancer detection. Our extensive quantitative experimental results on a large-scale cancer dataset, including 1,651 unique patients, validate that our approach can reduce human annotation efforts by at least 70\% while maintaining comparable cancer detection accuracy to competing fully supervised methods (AUC value 0.961 versus 0.966).

\end{abstract}

\section{Introduction}
\label{sec:intro}

In recent years, existing large-scale medical image datasets combined with deep learning techniques offer good feasibility for opportunistic patient disease or cancer detection at no additional radiation exposure to patients who underwent imaging exams for other routine purposes. Previous studies have shown that abdominal and chest CT scans provide significant values for incidental osteoporosis screening~\cite{jang2019opportunistic}, cardiovascular event prediction~\cite{pickhardt2020automated}, cancer detection \cite{cao2023large,chen2023cancerunit,yan2023liver}, etc. 
Compared with other screening tools like blood tests \cite{klein2021clinical} and endoscopic techniques \cite{chen2021effectiveness}, deep learning models via non-contrast CT have shown high performances in detecting different cancer types while mitigating the risk of over-diagnosis~\cite{cao2023large,ECScreening2022,yuan2023cluster}. Such non-contrast CT-based cancer detection ability may pave a path for effective large-scale population-based cancer screening \cite{kleeff2023ai}.

\begin{figure}[!t]
    \centering
    \includegraphics[width=0.8\linewidth]{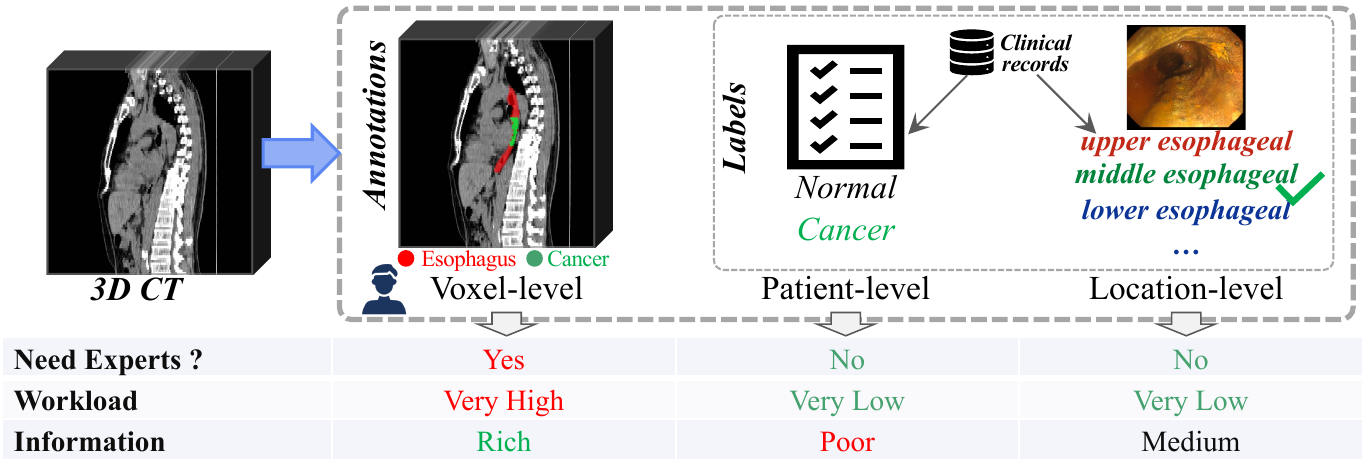}
    \caption{\small Illustration of available different types of annotations or labels for non-contrast CT-based Esophageal Cancer (EC) screening. Tumor annotating can be a labor-intensive process, if it involves carefully labeling the tumor mask on each axial CT slice. In contrast, clinical reports offer readily accessible patient-level and location-specific information, with the possibility of eliminating the need for time-consuming expert annotation, at least partially.}
    \label{fig:example}
  \end{figure}

Deep learning methods for disease and cancer detection~\cite{ECScreening2022,xia2021effective,zhu2019multi,pickhardt2020automated, zhang2021dodnet} typically rely on segmentation models that necessitate extensive clinical knowledge and voxel-level disease or tumor annotations. However, the high cost of per-voxel tumor annotations on CT imaging can make it unfeasible to acquire a large cancer screening dataset with all fully annotated tumors. Alternatively, clinical diagnosis reports, which are readily available through the Hospital Information System (HIS), contain valuable tumor-related information. 
For example, Fig.\ref{fig:example} showcases various annotations or labels essential for creating esophageal cancer (EC) screening model. Weak labels, including patient-level diagnosis labels and tumor location, are easily accessed from clinical documents like surgery or endoscopic reports. Conversely, precise tumor annotations demand that radiologists meticulously define tumor edges on successive CT slices—a process, which is notably laborious due to the low contrast between tumor and normal esophageal tissues on CT images. Furthermore, the lesion annotation process is highly subjective, often influenced by the annotators' individual preferences, leading to inter-observer variability that can introduce biases and impact the reliability of subsequent model training.
Given these challenges, developing a CT-based cancer screening model that achieves high accuracy with minimal manual labeling would be highly advantageous, which would greatly enhance data scalability and the generalizability of the system.

For efficient utilization of existing clinical data and reducing the annotation workload of radiologist experts, this work proposes a novel weakly semi-supervised learning (WSSL) method for medical image-based cancer screening. Specifically, we aim to combine a \textit{small} set of patients with voxel-level \textbf{tumor annotations} and a \textit{large} set of patients having diagnosis and tumor location labels from their \textbf{clinical reports} to train a non-contrast 3D CT-based cancer screening model. Following previous WSSL paradigm~\cite{chen2021points,zhang2022group,kim2023devil}, our WSSL cancer detection model consists of two steps: (i) training a tumor segmentation teacher network with fully annotated images to generate pseudo tumor masks for weakly-supervised images; (ii) training a cancer detection student network by using all images associated with both full annotations and pseudo annotations. This paradigm faces two problems: the teacher model is at risk of overfitting due to the scarcity of enough available tumor mask annotations, and the student model could be misdirected by inaccurate pseudo tumor annotations. Overcoming these challenges with the use of readily available weak labels from clinical reports would be highly beneficial.

However, introducing weak labels like tumor location through standard classification will make the network overfocus on the relationship between tumors and organs, neglecting other contextual information that potentially benefits cancer screening. This overemphasis on local information could diminish the network's generalization ability and robustness when dealing with few full annotations or noisy labels.
Therefore, this paper proposes a new text-guided learning scheme to extract and integrate knowledge from clinical reports into our WSSL 3D vision model. Specifically, we introduce the text encoder of a vision-language model (\eg, CLIP~\cite{radford2021learning}) to map text prompts of weak labels into high-dimensional latent space, so the optimization of weak labels can be carried out in the VLM latent space by measuring the similarity between projected image features and text features.
Our text-guided learning scheme can mine useful knowledge from weak labels in clinical reports to boost the efficacy of the WSSL cancer detection model. It mitigates the overfitting risk of the teacher network caused by the scarcity of tumor annotations by leveraging knowledge from the pre-trained VLM model. Besides, by infusing exact spatial tumor location details from reports, it can relieve the misleading effects of pseudo tumor masks on the cancer detection task. The main contributions of this work are summarized as follows:
\begin{itemize}
    \item We propose the first work for non-contrast CT-based cancer screening by constructing multi-modal class representations in a unified WSSL learning framework, achieving a good trade-off between minimizing expert annotation workloads and improving cancer detection performance.
    \item We present a novel text-guided mechanism to introduce knowledge of tumor location labels from clinical reports into a 3D CT-based vision model. Text representation is utilized to establish an optimization mechanism in the VLM latent space for weak supervision, beneficial to both tumor segmentation and cancer detection phases in WSSL.
    \item We extensively validate the effectiveness of our WSSL method on a large-scale esophageal cancer dataset of 1,651 unique patients. Our method with only 30\% tumor annotations can achieve equivalent performance (AUC=0.961) with the fully supervised model (AUC=0.966) on the test set.
\end{itemize}

\section{Related Work}
\label{sec:related}

\textbf{Weakly semi-supervised learning.} Weakly semi-supervised learning (WSSL) has been explored in the field of object detection~\cite{zhang2023vision, zhang2018learning,chen2021points,zhang2022group,ge2023point,zhang2023weakly}. Point DETR~\cite{chen2021points} combines small fully annotated images and many images labeled by points to train a DETR-like object detector. Group R-CNN~\cite{zhang2022group} and Point-Teaching~\cite{ge2023point} follow the same setting as Point DETR with further improved performance. Kim \etal~\cite{kim2023devil} extend this paradigm into the instance segmentation task, in which training data combines instance-level pixel supervision and point-level supervision. Recently, researchers have introduced the WSSL paradigm into applications like remotely sensed images~\cite{wang2023weakly} and lung ultrasound videos~\cite{li2023weakly,ouyang2023weakly}.
WSSL usually works in two steps: the first step is to train a teacher network with only fully annotated images and generate pseudo annotations for remaining images; the second step is designed to train a student network by using all images. WSSL can help to achieve a balance between performance and annotation costs by combining multiple types of supervision. It is important to point out that WSSL is more closely related to semi-supervised learning than to weakly-supervised learning (WSL). This distinction arises because WSL~\cite{ahn2018learning,jing2019coarse,jin2021isnet,ghiasi2022scaling} typically does not have any fully annotated data available, and its performance often exhibits a substantial gap compared to a fully supervised model. On the other hand, WSSL strives to achieve performance comparable to that of a fully supervised model.

\textbf{Vision-language learning in medical image.} Deep learning-based methods have been widely exploited to achieve promising results for organ segmentation and tumor detection~\cite{isensee2021nnu, cao2022swin, tang2022self, he2023swinunetr, chen2023cancerunit}. In recent years, vision-language models (VLMs) have attracted increasing attention which aim to jointly learn representation from joint visual and textual input~\cite{lu2019vilbert,radford2021learning,li2022blip,li2023blip,wu2023visual}. ViLBERT~\cite{lu2019vilbert} is a pre-trained model that can provide task-agnostic representation for images and texts. CLIP~\cite{radford2021learning} is a zero-shot learning model that learns a joint latent space for images and texts via aligned features of image-text pairs. Vision-language pre-trained models have been proven to be very efficient in various vision tasks~\cite{patashnik2021styleclip,wang2022clip,li2022ordinalclip,lin2023clip,xie2022clims,xu2023learning,liang2023crowdclip}. Recently, researchers have started to introduce vision-language mechanisms into the field of medical image analysis~\cite{lee2023text,wang2023pre,zhang2023large,Chen_2023_ICCV}, including pathology ~\cite{lu2023towards,huang2023visual,lu2023visual,yellapragada2023pathldm,lai2023clipath} and radiology tasks~\cite{wang2020doubleu,thawkar2023xraygpt,wang2022medclip,liu2023clip,liu2023clip2,Wu_2023_ICCV,li2023lvit}. For 3D CT images, Li \etal~\cite{li2023lvit} utilize BERTE to extract the embedding of the medical report and use it to guide the segmentation of the medical image. Liu \etal~\cite{liu2023clip,liu2023clip2} introduce CLIP model to the task of 3D CT-based organ segmentation and tumor detection and proposes a masked back-propagation method to resolve the partial labeling problem. Bosma \etal~\cite{bosma2023semisupervised} present a semi-supervised model that designs a post-processing strategy for pseudo masks informed according to lesion counts from prostate cancer reports.
In this paper, since 3D CT scans can not utilize the image encoder of CLIP, we introduce the text encoder of CLIP~\cite{radford2021learning} to make the weakly-supervised learning be optimized in the latent space of CLIP, which subsequently helps utilize a large amount of non-contrast CT scans with (only) tumor location labels available in pathology reports for training.

\section{Method}
\label{sec:method}

{\bf Problem Statement.} The overall goal of cancer screening is to predict whether the patient has cancer findings or not. As a rapid and non-invasive screening technique,   ``detection by segmentation'' scheme based CT cancer detection required voxel-level tumor annotations~\cite{ECScreening2022, xia2021effective,cao2023large}. Since a large percentage of solid tumors can exhibit low CT intensity contrast with normal soft tissues in non-contrast CT imaging, even experienced radiologists take a lot of effort on localizing and annotating tumor pixels. Radiologists usually would read and refer to patients' clinical reports (that indicate tumor findings and their anatomical locations in textual content even from a different imaging modality such as endoscopic exams for esophageal cancer patients), then semantically or mentally map these findings to CT modality space. Obtaining a very large-scale fully annotated cancer patient dataset purely by manual annotations can be very labor, time, and cost-intensive. Clinical reports already describe the tumor locations, which should be used as a form of weak supervision signal scalably to assist in training CT-based cancer screening models in the WSSL manner. In WSSL, we have a small set of CT scans with full supervision (\ie, both voxel-level tumor masks annotated by radiologists and patient-level weak labels from clinical reports)  $\mathcal{D}_{f} = \{(\mathbf{X}_{i}, \mathbf{M}_{i}), l_{i}, y_{i}\}_{i=1}^{N_{s}}$  and a large set of CT images with only weak labels $\mathcal{D}_{w} = \{(\mathbf{X}_{i}, l_{i}), y_{i}\}_{i=1}^{N_{w}}$. Here, $\mathbf{X}_{i}$ is the $i$-th non-contrast CT, $\mathbf{M}_{i}$ is the voxel-level tumor mask, $y_{i} \in \{0, 1\}$ is the cancer label, and $l_{i} \in [1, 2, ..., L]$ is the tumor location spatial index and $L$ denotes the number of discrete locations. In this paper, experiments are mainly conducted on esophageal cancer detection task, and the tumor location label is defined as one of the four anatomical \{\textit{upper, middle, lower, esophagogastric junction}\} portions of the esophagus or \{\textit{no}\} location if it is from a normal control, following the clinical practice standard. 

\subsection{Overall Framework}
Tumors usually occupy very tiny portions of the entire CT. Before the two-step training paradigm of WSSL, we first extract the esophagus from the original CT volume by organ ROI (region of interest) segmentation and extraction which is motivated by a similar step in~\cite{cao2023large}. We train a 3D U-Net to segment the esophagus, prune out unrelated backgrounds, and add a margin of $32\times 32\times 4$ voxels to obtain the final esophagus ROI. The 3D U-Net is trained on the SegTHOR dataset~\cite{lambert2020segthor} to output the esophagus region given a CT scan input. A self-learning process~\cite{zhang2018self} is used to refine the final esophagus masks on our training set, and our model can obtain a Dice score of 0.85 on the test set of SegTHOR dataset. \textit{Notably, all models in this paper are trained after the esophagus segmentation process.}

\begin{figure}[t]
    \centering
    \includegraphics[width=1\linewidth]{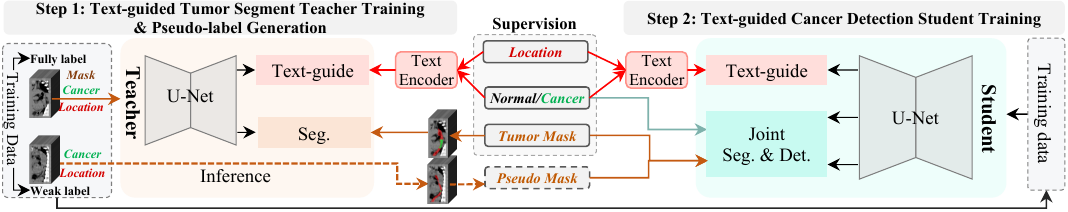}
    \caption{\small  Overall framework of our text-guided weakly semi-supervised cancer detection framework. \textbf{Step 1:} Training a tumor segmentation teacher network on a subset of fully annotated images (\eg 30\%) for detailed cancer characteristics. Then, leverage this network to create pseudo tumor masks for the remaining weakly-supervised images (\eg 70\%) that only have clinical reports.
    \textbf{Step 2:} Using all training images to train a cancer detection student network. Notably: our text-guided learning scheme, which utilizes weak labels, serves dual purposes: it mitigates the overfitting associated with limited training data in Step 1, and it compensates for the imprecision of pseudo tumor masks in Step 2.
    } 
    \label{fig:framework}
\end{figure}

After the ROI extraction, the overall framework of our weakly semi-supervised cancer screening framework is illustrated in Fig.~\ref{fig:framework}, which consists of two steps. 

\noindent\textbf{Step 1. Text-guided segmentation teacher.}
In this step, CT scans with fully annotated voxel masks are used to train a segmentation teacher network. In our setting, patients with full annotations are from a small portion of all training data (\eg, 30\%).
The teacher network consists of a segmentation network and our text-guided sub-network (details in Section~\ref{sec:text-guided}). Text-guided sub-network introduces both the location and diagnosis labels and is optimized by the text-guided loss. 
The overall loss function of the teacher network is the combination of segmentation loss and text-guided loss:
\begin{equation}
   \mathcal{L}_{teacher} = \mathcal{L}_{seg} + \lambda\mathcal{L}_{text},
\end{equation}
where $\mathcal{L}_{seg}$ is the segmentation loss which combines cross-entropy loss and Dice loss. $\mathcal{L}_{text}$ is the loss for our text-guided learning scheme, and $\lambda$ is set as 0.01. After training, the teacher network is used to generate pseudo tumor masks for the rest of the CT scans (\ie, 70\% patients). In this step, our text-guided learning can mitigate the overfitting of the segmentation network associated with limited training data (see Table~\ref{tab:seg}).

\noindent\textbf{Step 2. Text-guided cancer detection student.}
In this step, we use all training CT scans to train a cancer detection student network. As shown in Fig.~\ref{fig:method}, the text-guided cancer detection student consists of a joint tumor segment and cancer detection network (details in Section~\ref{sec:joint}), and our Text-guided weak information mining network (details in Section~\ref{sec:text-guided}). 
The text-guided network introduces location and diagnosis labels from reports to compensate for the imprecision of pseudo tumor masks. The overall loss function of the student is:
\begin{equation}
    \mathcal{L}_{student} = \mathcal{L}_{joint} + \alpha\mathcal{L}_{text},
\end{equation}
where $\mathcal{L}_{joint}$ is used for the joint segmentation and detection framework (Eq.~\ref{eq:loss_joint}); and our text-guided $\mathcal{L}_{text}$ loss (Eq.~\ref{eq:loss_text}) will be described in details in Section~\ref{sec:text-guided}.

Compared to previous fully supervised cancer screening methods~\cite{xia2021effective, zhang2021dodnet}, our weakly semi-supervised cancer screening framework can efficiently utilize large-scale clinical reports that often already exist to improve the performance and accuracy of CT-based cancer detection and screening, reducing the otherwise required annotation workloads by radiologists.

\subsection{Joint Tumor Segmentation \& Cancer Detection}
\label{sec:joint}
In this section, we describe the joint learning framework for tumor segmentation and cancer detection. 
We provide a brief description of each part as follows.

\begin{figure}[!t]
    \centering
    \includegraphics[width=1.0\linewidth]{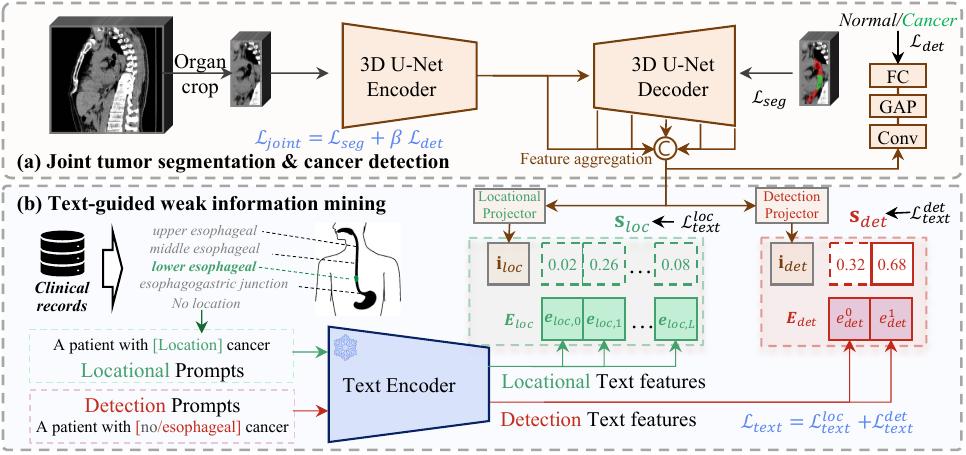}
    \caption{Illustration of our proposed text-guided cancer detection framework. \textbf{(a)} A joint learning framework of tumor segmentation and cancer detection, multi-scale features from the decoder of a segmentation network are aggregated together and fed to a classification head to obtain the final screening results. \textbf{(b)} A text-guided learning scheme that can mine information from texts as weak labels in clinical records, including diagnostic labels and tumor location labels.}
    \label{fig:method}
  \end{figure}
  
\textbf{Backbone and feature aggregation.} As shown in Fig.~\ref{fig:method} (a), given a 3D ROI CT, we feed it into a 3D U-Net-like backbone and obtain multi-scale feature maps $\mathbf{F} = \{\mathbf{F}_{0}, \mathbf{F}_{1}, ..., \mathbf{F}_{S}\}$ from the image decoder, where $S$ is the number of stages, and $\mathbf{F}_{i}$ is the feature map of the $i$-th stage. Then, we standardize the spatial dimensions of all feature maps from the decoders to a uniform spatial size using trilinear interpolation. Subsequently, we concatenate these maps by aligning them along their depth dimension, forming an aggregated feature representation that serves as the input for the subsequent cancer detection task.

\textbf{Tumor segmentation and cancer detection.} The segmentation head takes the last feature map as input and outputs the segmentation result. The cancer detection head takes the aggregated feature map as input and outputs the probability of the patient having cancer. It consists of one 3D convolution layer with kernel size $3\times 3\times 3$, one global average pooling layer, and one fully connected layer, as illustrated in Fig.~\ref{fig:method} (a). The overall loss of the joint learning network is:
\begin{equation}
    \mathcal{L}_{joint} = \mathcal{L}_{seg} + \beta\mathcal{L}_{det}.
    \label{eq:loss_joint}
\end{equation}
where $\mathcal{L}_{seg}$ is the loss for the 3D U-Net segmentation network; $\mathcal{L}_{det}$ is cross-entropy loss for the cancer detection task and $\beta$ is its weight.

\subsection{Text-Guided Weak Information Mining}
\label{sec:text-guided}

Voxel-level annotations are needed in recent cancer detection methods ~\cite{zhang2021dodnet,ECScreening2022,zhu2019multi} but it can be limited by the heavy annotation workload required by radiology experts in the practice. This work aims to utilize weak clinical labels in diagnosis reports to achieve good cancer detection results without needing excessive amounts of human annotations. 
In this paper, to adequately utilize the report information, we propose a new text-guided learning scheme as follows. As depicted in Fig.~\ref{fig:method}(b), the main motivation is to employ the pre-trained text encoder to establish a structured high-dimensional latent space, so that our trained model can learn the spatial relationship between the tumor and the organ in an implicit way, more feasible for network optimization.

\textbf{Architecture}. Based on the aggregated multi-scale features, two projectors are introduced to generate two image features to learn information from text cues: (i) one projector generates an image feature vector $\mathbf{i}_{det} \in \mathbb{R}^{1\times D}$, which is used to learn from cancer diagnostic texts; (ii) another projector generates an image feature vector $\mathbf{i}_{loc} \in \mathbb{R}^{1\times D}$, which is used to learn from tumor location texts. Here, $D$ is the dimension of the CLIP latent space~\cite{radford2021learning}. These two heads have the same architecture which consists of one 3D convolution layer with kernel size $3\times 3\times 3$ and the channel number $D$, one global average pooling layer, and one fully connected layer. 

\textbf{Prompt design.} In our task, besides the diagnosis labels, we also consider the tumor location information in our prompt design. Therefore, we design a prompt template to utilize the report information: ``A patient with {\{Label\} cancer''. The \{Label\} is confirmed by pathology reports, including two kinds of diagnostic labels and $L$ tumor location labels. The diagnostic labels indicate whether a patient is normal or abnormal (\eg, having ``esophageal'' cancer).
$L$ tumor location labels are defined as one of the five locations, \{\textit{upper, middle, lower, esophagogastric junction}\} portions of the esophagus, plus \{\textit{no}\} location, following the clinical practice standard. Feeding these two types of prompts into the CLIP text encoder, we first obtain detection text features $\mathbf{E}_{det} =[\mathbf{e}_{det}^{0}$, $\mathbf{e}_{det}^{1}]$. Then, tumor location labels include $L$ tumor location names (\eg, ``upper esophageal'') with one background name (\ie, ``no''), and we use the same prompt template to obtain $L+1$ locational text features $\mathbf{E}_{loc} = [\mathbf{e}_{loc, 0}, \mathbf{e}_{loc, 1}, ..., \mathbf{e}_{loc, L}]$. All text features are of the same dimension $1\times D$. 

\textbf{Text-guided optimization.} In the learning of tumor position information, we use all tumor location text features $\mathbf{E}_{loc}$ to build a structured output space for the locational information. Given the projected image feature vector $\mathbf{i}_{loc}$ from the locational projector, we calculate the similarity between $\mathbf{i}_{loc}$ and $\mathbf{E}_{loc}$ and obtain the similarity vector $\mathbf{s}_{loc} = [s_{loc, 0}, s_{loc, 1}, ..., s_{loc, L}]$:
\begin{equation}
    s_{loc, i} = \mathbf{i}_{loc} \cdot \mathbf{e}_{loc, i}^\mathsf{T},
    \label{eq:similarity}
\end{equation}
this similarity vector $s_{loc, i}$ is regarded as the predicted tumor location result and can be optimized by cross-entropy loss. Following the training process of CLIP~\cite{radford2021learning}, a softmax operation with a learnable temperature parameter $T_{loc}$ is introduced before the cross-entropy loss: for the $i$-th similarity score $s_{loc, i}$, we have $p_{loc, i} = {\rm exp}(s_{loc,i}/T_{loc})/\sum_{j=0}^{L}{\rm exp}(s_{loc,j}/T_{loc})$. The cross-entropy loss for the tumor location head is:
\begin{equation}
    \mathcal{L}_{text}^{loc} = -\sum_{i=0}^{L}\hat{p}_{loc, i}\log{p_{loc,i}},
    \label{eq:loss_loc}
\end{equation}
where $\hat{p}_{loc,i} \in \{0, 1\}$ is the ground-truth probability of the tumor location label.

Similar to the learning process of the location head, we use two detection features to build a structured output space for diagnostic labels, \ie, $\mathbf{E}_{det} =[\mathbf{e}_{det}^{0}$, $\mathbf{e}_{det}^{1}]$. The similarity vector $\mathbf{s}_{det}$ is calculated in a similar way as the Eq.~\ref{eq:similarity}. The softmax with temperature operation is also used to obtain the probability of the patient label: $p_{det, i} = {\rm exp}(s_{det,i}/T_{det})/\sum_{j=0}^{1}{\rm exp}(s_{det,j}/T_{det})$. The cross-entropy loss for the text-guided detection head is:
\begin{equation}
    \mathcal{L}_{text}^{det} = -\sum_{i=0}^{1}\hat{p}_{det, i}\log{p_{det,i}},
    \label{eq:loss_pat}
\end{equation}
where $\hat{p}_{det,i} \in \{0, 1\}$ denotes whether the patient has cancer or not. The overall loss function of the text-guided learning scheme is:
\begin{equation}
    \mathcal{L}_{text} = \mathcal{L}_{text}^{loc} + \mathcal{L}_{text}^{det}.
    \label{eq:loss_text}
\end{equation}

Our text-guided mechanism leverages the weak labels in clinical reports to extract valuable diagnosis information, benefiting both phases of WSSL: (i) During the training of the tumor segmentation teacher network, our technique utilizes the extensive knowledge of a pre-trained VLM to improve the network's generalization capabilities despite the small amount of annotated data. (ii) When training the student network, our method corrects errors in the process of pseudo tumor masks by incorporating precise tumor location information from the reports, thereby enhancing or enforcing the overall accuracy of the diagnosis.

\section{Experiments}
\label{sec:experiments}
\vspace{-0.3cm}
\subsection{Experimental Setting}

\textbf{Dataset and ground truth.} In this paper, we collect a large-scale esophageal cancer screening dataset of 1,651 patients from one cancer medical center with IRB approval. 980 esophageal cancer patients have both 3D non-contrast CT scans and associated surgery reports, and 671 patients with negative diagnoses determined by the follow-up clinical records are used as normal controls. To acquire non-contrast tumor masks, an experienced radiologist (10 years esophageal cancer specialist) manually annotates tumors on the contrast-enhanced CT by mentally referring to the corresponding clinical reports (\ie, endoscopic, and surgery reports) as well. Then a robust registration algorithm~\cite{deeds} is used to transfer annotations from the manually annotated contrast-enhanced CT to non-contrast CT scans. This dataset is further split into 1,056 training, 264 validation, and 331 test sets, and we keep roughly similar cancer versus normal patient ratios in each split. We simulate the WSSL setting by splitting 1,056 training patients into two parts, \eg 30\% patients (317) with both annotations and weak labels, and the rest 70\% patients (739) with only weak labels for training. The best model on the validation is set to inference the test set.

\textbf{Metrics and evaluation.} For tumor segmentation in the teacher model, the Dice score is tested. For tumor location prediction, we calculate the accuracy (ACC). For cancer detection, AUC (Area under Curve), sensitivity, and specificity scores are computed. For sensitivity and specificity, the cutoff is set as 0.5. The best teacher segmentation model in the validation set is selected for inference in the simulated unlabeled set (\eg 70\% of 1,056 training patients) and the test set. For student model, we select the best model for cancer detection in the validation set and report results in the test set.

\textbf{Implementation details.} Each CT volume is resampled into $0.7 \times0.7 \times 5$ mm spacing and normalized into zero mean and unit variance. After the ROI extraction, the esophagus CT subvolume is then resized into the shape of $96 \times 96 \times 64$. In the training process, we use data augmentation techniques such as random rotation, random scaling, random flipping, and random intensity scaling for robustness. We use the base SwinUnetV2~\cite{he2023swinunetr} as the backbone for our model. The text encoder is $ViT-L/14@336$px from CLIP~\cite{radford2021learning}, and the length of the text feature vector is 768. We use the Adam optimizer with a warm-up cosine scheduler of 50 epochs, the initial learning rate is set to 0.0005, and the weight decay is set to 0.0001. The batch size is set to 6. The loss weight $\alpha$ and $\beta$ are set as 0.01 and 0.1 according to ablation study on the validation set, respectively. The model is trained on 1 NVIDIA V100 GPU with 32GB memory. The framework is implemented using MONAI 1.13.0~\cite{cardoso2022monai} and PyTorch 1.12.0~\cite{paszke2019pytorch}.

\textbf{Comparison baselines.} We compare our method to baselines trained in different experimental settings. (i) Pure classification models (with only weak supervisions) with SwinUNetr~\cite{tang2022self} and SwinUNetrV2~\cite{he2023swinunetr} backbones. (ii) ``Detection by segmentation''-based EC screening model AxialUNet~\cite{ECScreening2022} and the most recent state-of-the-art cancer detection method PANDA~\cite{cao2023large} with different backbones. Those methods are trained in two settings by using 100\% or 30\% voxel-level tumor annotations, respectively. (iii) PANDA~\cite{cao2023large} with SwinUNetrV2~\cite{he2023swinunetr} backbone in the same WSSL setting (30\% fully supervisions + 70\% weak supervisions), which can be seen as a strong baseline without text-guided learning. 

\begin{table}[!t]
  \centering
  \scriptsize
  \caption{\textbf{Experimental results of cancer detection.} \textbf{(a)} Models trained by 100\% weak labels. \textbf{(b)} Models trained by 100\% fully-supervised data. \textbf{(c)} Only 30\% of data with full supervisions are used in training. \textbf{(d)} Weakly semi-supervised setting: training data consists of 30\% data with full supervisions and 70\% data with weak labels from clinical reports.}
  \setlength{\tabcolsep}{1.5mm}{
  \begin{tabular}{c|l|ccc|ccc}
    \hline
          &       & \multicolumn{3}{c|}{Val} & \multicolumn{3}{c}{Test} \bigstrut[t]\\
    Setting & Method & AUC   & Sens. & Spec. & AUC   & Sens. & Spec. \bigstrut[b]\\
    \hline
    \textbf{(a) } & SwinUNetr & 0.911  & 0.868  & 0.714  & 0.907  & 0.888  & 0.733  \bigstrut[t]\\
    100\% weak-supervised & SwinUNetrV2 & 0.919  & 0.882  & 0.795  & 0.902  & 0.852  & 0.756  \bigstrut[b]\\
    \hline
          & AxialUNet & 0.951  & 0.901  & 0.893  & 0.938  & 0.893  & 0.793  \bigstrut[t]\\
    \textbf{(b)} & PANDA-Original & 0.962  & 0.895  & 0.902  & 0.954  & 0.944  & 0.852  \\
    100\% fully-supervised & PANDA-SwinUNetr & 0.967  & 0.901  & 0.902  & 0.959  & 0.908  & 0.904  \\
          & PANDA-SwinUNetrV2 & \textbf{0.973} & 0.934  & 0.929  & \textbf{0.966} & 0.939  & 0.874  \bigstrut[b]\\
    \hline
          & AxialUNet & 0.905  & 0.757  & 0.929  & 0.912  & 0.699  & 0.889  \bigstrut[t]\\
          & PANDA-Original & 0.930  & 0.732  & 0.921  & 0.917  & 0.750  & 0.903  \\
    \textbf{(c)} & PANDA-SwinUNetr & 0.933  & 0.743  & 0.938  & 0.913  & 0.765  & 0.889  \\
    30\% fully-supervised & PANDA-SwinUNetrV2 & 0.945  & 0.743  & 0.964  & 0.923  & 0.760  & 0.904  \\
          & Ours-SwinUNetrV2 & \textbf{0.952} & 0.862  & 0.946  & \textbf{0.942} & 0.791  & 0.926  \bigstrut[b]\\
    \hline
    \textbf{(d)} & PANDA-SwinUNetrV2 & 0.953  & 0.895  & 0.938  & 0.946  & 0.867  & 0.896  \bigstrut[t]\\
    30\%+70\% WSSL & Ours-SwinUNetrV2 & \textbf{0.966} & 0.888  & 0.982  & \textbf{0.961} & 0.903  & 0.904  \bigstrut[b]\\
    \hline
    \end{tabular}%
    }
  \label{tab:results}
\end{table}

\subsection{Experimental Results} 
From Table~\ref{tab:results}, we provide cancer detection results across different settings. As shown in Table~\ref{tab:results}(a), we first describe results from two classifiers with only weak diagnosis labels for training. We could see pure classification models cannot achieve highly satisfied AUCs in both validation and test sets which implies the difficulties for cancer detection without using any tumor annotations. PANDA models~\cite{cao2023large} with only 30\% full supervisions achieve better results than two classifiers, indicating the importance of tumor annotations in cancer detection. In Table~\ref{tab:results}(c), among all models under 30\% setting, our final model achieves the best AUCs on both sets which demonstrates the effectiveness of the proposed text-guided learning.

The 100\% fully-supervised setting can be seen as the performance upper bound of the WSSL setting. From Table \ref{tab:results}(b), the best model is PANDA~\cite{cao2023large} with SwinUNetrV2~\cite{he2023swinunetr} backbone which achieves the AUC of 0.973 and 0.966 on the validation and test set, respectively. For the WSSL settings in Table \ref{tab:results}(d), we compare our method with this very strong baseline (PANDA-SwinUNetrV2).  Our method achieves an AUC of 0.961 on the test set, which is higher than PANDA-SwinUNetrV2 (AUC 0.946) under the same WSSL setting. In Fig.~\ref{fig:ROC}, we provide ROC curves of the proposed method and PANDA in different settings. The DeLong test is used to test statistical significance. Our method using only 30\% tumor annotations is statistically equivalent to the PANDA model in 100\% fully-supervised setting ($p=0.39$), which demonstrates our model is comparable to that of the fully-supervised method with no significant difference.
More results using different ratios of annotations can be seen in the Table \ref{tab:ratio2} and Fig.\ref{fig:roc_ratio}.

\subsection{Ablation Study and Model Analysis}
  \label{sec:ablation}

\textbf{Effects of text-guided learning in tumor segmentation teacher.}
Table~\ref{tab:seg} presents the effectiveness of the proposed text-guided learning in the step 1 tumor segmentation teacher model. The pure SwinUNetrV2~\cite{he2023swinunetr} model and our proposed method are trained on 30\% fully annotated data, and evaluated on the rest 70\% of training data, validation, and test set, respectively. After introducing the text-guide module, tumor segmentation can be improved from the pure segmentation model by 4.1\%-8.6\% of Dice scores for tumors in esophagus and 1.5\%-7.2\% for tumors in EGJ. Fig.\ref{fig:show_teacher} shows two early-stage cancer examples, our model could achieve better pseudo tumor masks than pure segmentation model SwinUNetrV2~\cite{he2023swinunetr}. The teacher model with text-guided learning could provide more reliable pseudo masks for the step 2 cancer detection student model.

\textbf{Analysis about joint learning framework without text guidance.}
Table~\ref{tab:joint} shows an ablation study in 30\% fully supervised data to analyze the joint learning framework in different settings, where cancer diagnostic and tumor location labels are utilized via standard classification paradigms. Table~\ref{tab:joint}(a) and (b) show classification methods cannot provide accurate results without using annotations, which indicate difficulties for cancer detection or tumor localization task in a pure weakly-supervised manner. From Table~\ref{tab:joint}(c) and (d), the joint learning with segmentation together can improve the localization accuracy from 0.613 to 0.687 in the term of ACC, and the cancer detection performance from 0.885 to 0.945 measured by AUC. However, results from Table~\ref{tab:joint}(e) indicate performances become worse in the multi-task learning for both tumor localization and cancer detection together with segmentation. The main reason is that such a joint model may neglect contextual information about cancer when over-focusing on the relationship between tumors and organs. Finally, we use the joint learning framework (c) in the student model.

\begin{figure}[!t]
\begin{minipage}[!b]{0.52\textwidth}
\makeatletter\def\@captype{figure}
\centering
    \includegraphics[width=1.0\linewidth]{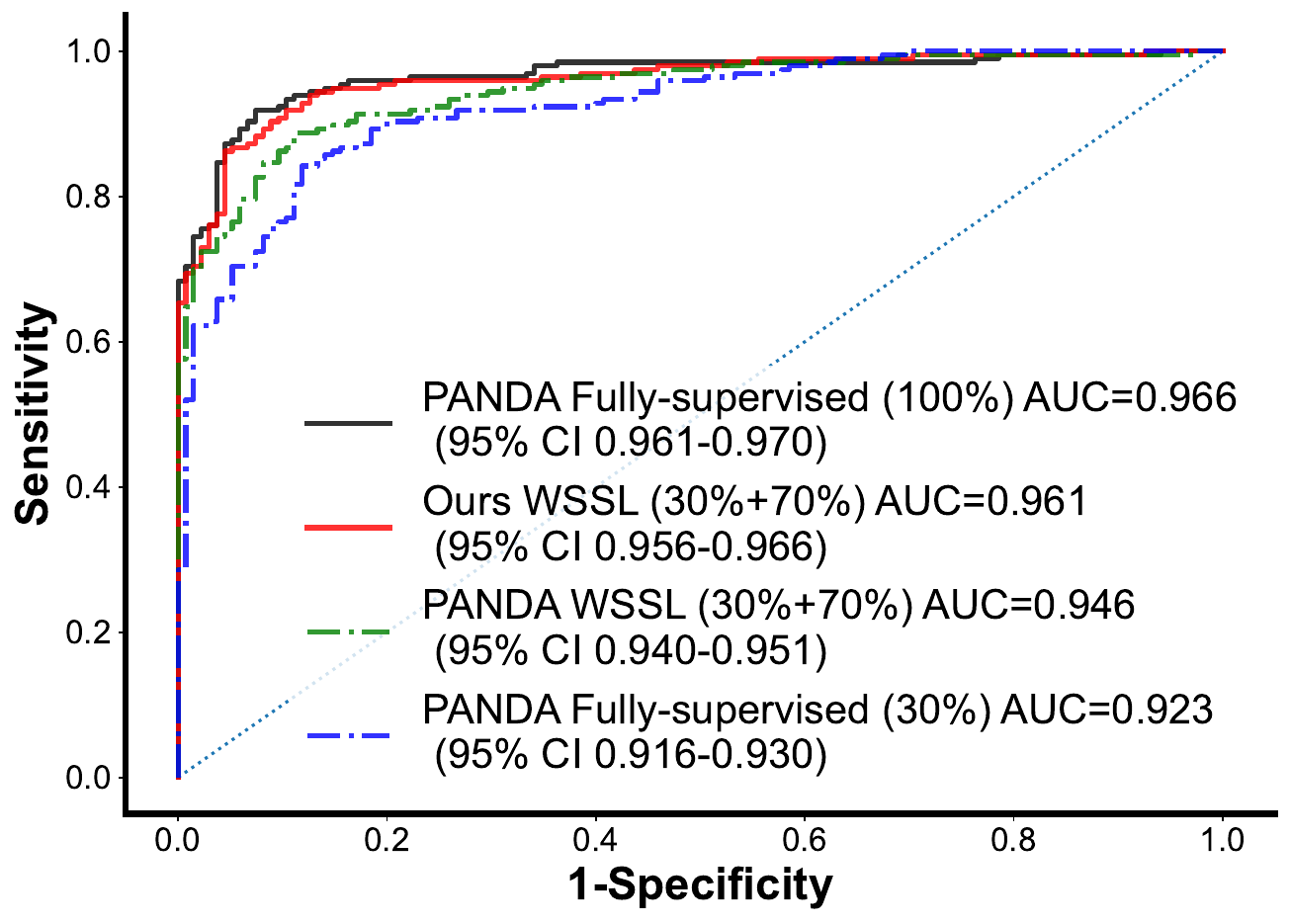}
    \caption{ROC curves of fully-supervised and WSSL model settings.}
    \label{fig:ROC}
\end{minipage}  
\quad 
\begin{minipage}[!b]{0.44\textwidth}
\makeatletter\def\@captype{table}
\centering
\scriptsize
\caption{Ablation study on step 1 tumor segmentation teacher model. We report Dice scores of tumors in  Esophagus (ESO) and Esophagogastric Junction (EGJ), respectively.}
  \setlength{\tabcolsep}{0.5mm}{
  \renewcommand{\arraystretch}{1.1}{
  \begin{tabular}{c|l|cc}
    \hline
    Set   & Teacher & ESO   & EGJ \bigstrut\\
    \hline
    70\% Weak  & Pure Seg. & 0.415  & 0.641  \bigstrut[t]\\
     Train & \quad+Text-guide & \textbf{0.438} & \textbf{0.650} \bigstrut[b]\\
    \hline
    \multirow{2}[2]{*}{Val} & Pure Seg. & 0.383  & 0.622  \bigstrut[t]\\
          & \quad+Text-guide & \textbf{0.416} & \textbf{0.667} \bigstrut[b]\\
    \hline
    \multirow{2}[2]{*}{Test} & Pure Seg. & 0.417  & 0.597  \bigstrut[t]\\
          & \quad+Text-guide & \textbf{0.434} & \textbf{0.619} \bigstrut[b]\\
    \hline
    \end{tabular}%
    }}
  \label{tab:seg}%
\end{minipage}
\end{figure}

\begin{figure}[!t]
    \centering
    \includegraphics[width=1\linewidth]{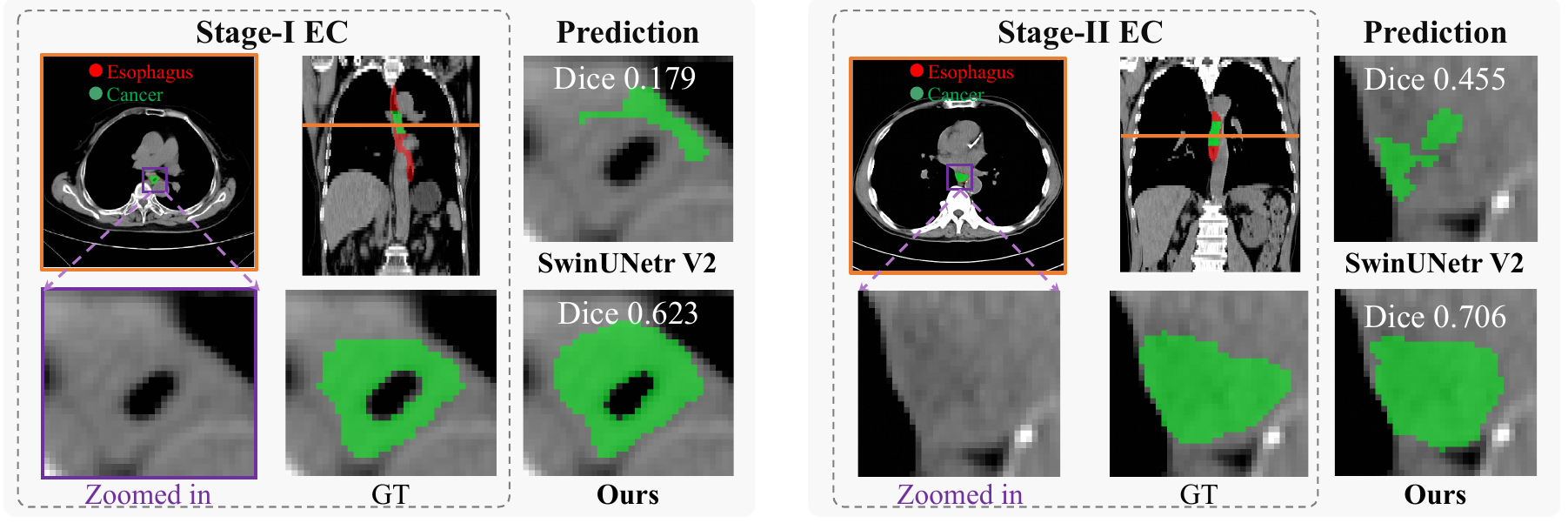}
    \caption{Visualization of pseudo tumor masks (Dice) in step 1 on 70\% weakly-supervised images. We provide patients with TNM stage I (left) and II (right), respectively. Our model can predict better pseudo tumor masks than the pure segmentation network. 
    }
    \label{fig:show_teacher}
  \end{figure}

  \begin{figure}[!t]
\begin{minipage}[!b]{0.52\textwidth}
\makeatletter\def\@captype{table}
\centering
\scriptsize
  \caption{Analysis on the joint learning network in student model. We report the accuracy of tumor localization (Loc.) and AUC for cancer detection (Det.) on validation set.}
  \setlength{\tabcolsep}{1.2mm}{
  \renewcommand{\arraystretch}{1}{
    \begin{tabular}{c|ccc|cc}
    \hline
          & Seg.  & Det.  & Loc.  & Det.(AUC)   &  Loc.(ACC) \bigstrut[b]\\
    \hline
    (a) &       & $\checkmark$ &       & 0.885  & - \bigstrut[t]\\
    (b)   &       &       & $\checkmark$ & -     & 0.613  \bigstrut[b]\\
    \hline
    (c) & $\checkmark$ & $\checkmark$ &       & \textbf{0.945 } & - \bigstrut[t]\\
    (d)   & $\checkmark$ &       & $\checkmark$ & -     & \textbf{0.687 } \\
    (e)   & $\checkmark$ & $\checkmark$ & $\checkmark$ & 0.932  & 0.643  \bigstrut[b]\\
    \hline
    \end{tabular}%
    }}
  \label{tab:joint}%
\end{minipage}  
\quad 
\begin{minipage}[!b]{0.44\textwidth}
\makeatletter\def\@captype{table}
\centering
\scriptsize
  \caption{Ablation study on text-guided learning in the student model. We report AUCs for cancer detection on both validation and test set.}
  \setlength{\tabcolsep}{0.3mm}{
  \renewcommand{\arraystretch}{1.5}{
    \begin{tabular}{l|c|c}
    \hline
    Student & Val   & Test \bigstrut\\
    \hline
    Joint Seg.\&Det. & 0.945  & 0.923  \bigstrut[t]\\
    \quad+Text-guide Det. & 0.947  & 0.934  \\
    \quad+Text-guide Det.\&Loc. & \textbf{0.952} & \textbf{0.942} \bigstrut[b]\\
    \hline
    \end{tabular}%
    }}
  \label{tab:text}%
\end{minipage}
\end{figure}

\textbf{Effects of weak lables in text-guided learning.}
In Table~\ref{tab:text}, we analyze how each type of text prompts affects the cancer detection performance. Models are trained on 30\% fully supervised images, and we report the cancer detection results on both validation and test set. We take the joint segmentation and detection network as the base model, which can obtain a 0.923 AUC value on the test set. By introducing only detection prompts, our text-guided mechanism can improve the cancer detection performance to AUC of 0.934. After incorporating tumor location prompts as well, our method can further improve the AUC to 0.942, which demonstrates that the text-guided method can effectively integrate different kinds of weak labels derived from the clinical reports. This is because the use of tumor location-aware and detection prompts is carried out in the latent space of the VLM model that may better perceive the intra-class diversities and inter-class contexts to benefit both cancer localization and detection tasks.

\textbf{Effects of $\alpha$ and $\beta$.} Table \ref{tab:alpha}-\ref{tab:beta} shows the effects of loss weights in the proposed framework on the validation set. All experiments are trained with 30\% fully supervised data. Table \ref{tab:alpha} presents the effects of the text-guided loss $\mathcal{L}_{text}$. We evaluate the AUC value for cancer detection from two dimensions: the classification head in the joint learning network, and the similarity to the detection text features in the text-guided network. Both classification outcomes show parallel variations, both of them achieve the highest AUC values when $\alpha$ is set to 0.01. When $\alpha$ is excessively high (\eg, 1), the effectiveness of text-guided learning suffers, leading to inferior results in the joint learning network with an AUC of 0.939.
Table \ref{tab:beta} shows when only the joint learning mechanism is used, the best cancer detection is with $\beta$ as 0.1.

\begin{table}[!t]
  \centering
  \scriptsize
  \caption{Ablation study on validation set about loss weight $\alpha$ of $\mathcal{L}_{text}$. ``Joint'' denotes results from the joint learning network, and ``Text'' denotes results from our text-guided detection branch.}
  \vspace{-0.2cm}
  \setlength{\tabcolsep}{4mm}{
    \begin{tabular}{c|ccccc}
    \hline
    $\alpha$ & 0.0001 & 0.001 & 0.01  & 0.1   & 1 \bigstrut\\
    \hline
    AUC (Text) & 0.921  & 0.942  & \textbf{0.951} & 0.932  & 0.913    \\
    AUC (Joint) & 0.942  & 0.947  & \textbf{0.952} & 0.944  & 0.939 \bigstrut[b]\\
    \hline
    \end{tabular}%
    }
  \label{tab:alpha}%
\end{table}%

\begin{table}[!t]
\vspace{-0.2cm}
  \centering
  \scriptsize
  \caption{Ablation study on validation set about loss weight $\beta$ of $\mathcal{L}_{det}$.}
  \vspace{-0.2cm}
  \setlength{\tabcolsep}{3mm}{
    \begin{tabular}{c|ccccccc}
    \hline
    $\beta$ & 0.01  & 0.05  & 0.1   & 0.15  & 0.2   & 0.25  & 1 \\
    \hline
    AUC   & 0.926 & 0.940 & \textbf{0.945} & 0.943 & 0.936 & 0.937 & 0.931 \\
    \hline
    \end{tabular}%
    }
  \label{tab:beta}%
\end{table}%

\textbf{Visualizations of step 2 student model in the WSSL setting.} In Fig.~\ref{fig:show}, we provide visualization results of WSSL models on two patient subjects with cancer TNM stage I and II, respectively. Stage I is the early stage, in which the tumor is of low contrast with normal tissue and is very difficult to locate, we can find that our method can predict a small part of the tumor and correctly predict the cancer with a probability of 0.987, while the comparison PANDA-SwinUNetrV2 method can not detect the cancer in this CT. Stage II patient is of medium severity, for this case, our method can predict a considerable portion of the tumor and give a more confident detection probability than PANDA baseline predictions. Our model correctly predicts tumor locations for above two cases.

  \begin{figure*}[!t]
    \centering
    \includegraphics[width=1\linewidth]{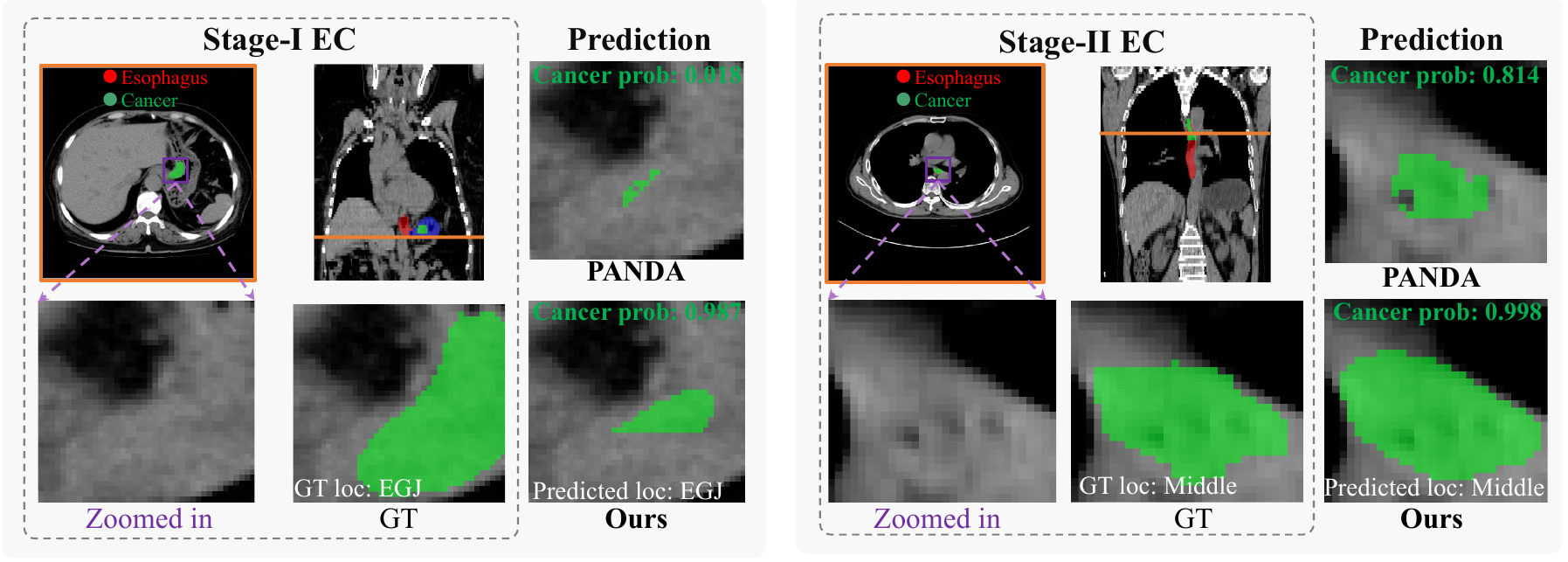}
    \caption{Visualization of WSSL step 2 cancer detection model on the test set. We provide patients with TNM stage I (left) and II (right). Our model can predict higher cancer probability while accurately providing its location.
    }
    \label{fig:show}
  \end{figure*}

\section{Conclusion}
\label{sec:conclusion}

In this paper, we propose a novel text-guided learning method for effective cancer screening that exploits and leverages clinically already existent large amounts of clinical reports to boost the medical image-based detection models that usually would require pixel-wise tumor mask annotations. We utilize patient diagnosis labels and tumor location descriptions in the reports to optimize the vision models, and a text-guided mechanism is proposed to incorporate the report information into a weakly semi-supervised framework. Specifically, the text encoder of CLIP is used to extract text features of report labels. These text features come from the latent space of the CLIP can be used to construct a structured output space, in which cancer and location information are predicted by similarity matching between two predicted features and the generated CLIP features. Our text-guided learning schemes can be easily integrated into the existing vision models. The report information can subsequently be employed to optimize/improve the vision models in a weakly semi-supervised manner. 
Our proposed approach is extensively evaluated on the esophageal cancer screening task, and quantitative results have shown that we can achieve comparable performance with the fully supervised method with only 30\% of the full annotations. Our method can be straightforwardly extended to screening tasks for other types of cancer, which may be a future direction of this work.

\noindent \textbf{Limitations:} Although tumor location is crucial, clinical reports also contain other potentially useful information that are not well investigated in this study. More comprehensive utilization of clinically relevant information in reports to train scalable medical AI systems needs to be explored in future work.

\bibliographystyle{splncs04}
\bibliography{main}

\newpage
\appendix

\section{Trade-off of annotation budgets and performance.}
\label{sec:experiments}

\noindent \textbf{A1. Step 1 tumor segmentation under different data proportions.}
In our exploration of the step-1 tumor segmentation performance in Table~\ref{tab:ratio1}, we utilized a teacher network that was trained under varying proportions of fully-supervised datasets—specifically 10\%, 20\%, 30\%, and 40\%. Performance is evaluated using the complementary 90\%, 80\%, 70\%, and 60\% portions of weakly-supervised training data, respectively. Dice for tumors in the esophagus (ESO) and esophagogastric junction (EGJ) are reported.
`Pure Seg.' represents a pure SwinUNetrV2~\cite{he2023swinunetr} model. With an increase in fully-supervised data from 10\% to 40\%, we observed a consistent growth in the Dice scores for both ESO and EGJ, indicating a positive correlation between the amount of fully-supervised data and segmentation accuracy.
Our proposed method, indicated by `+ Text-guided', incorporates text-guided learning into the segmentation process. The text-guided method consistently outperformed the pure segmentation across all proportions of fully-supervised data, as evidenced by the higher Dice scores—reaching 0.303 and 482 for ESO and EGJ respectively with just 10\% of fully-supervised data, and further improving to 0.463 and 0.667 with 40\% fully-supervised training.

\begin{table}[!h]
\vspace{-0.7cm}
  \centering
  \begin{minipage}{0.55\linewidth} 
    \centering
    \footnotesize
    \setlength{\tabcolsep}{0.9mm}{
      \renewcommand{\arraystretch}{0.6}{
    \begin{tabular}{c|l|cc}
    \hline
    Label Types & Method & ESO   & EGJ \bigstrut\\
    \hline
    \multirow{2}[2]{*}{$\mathcal{F}$ 10\%} & Pure Seg. & 0.286  & 0.437  \bigstrut[t]\\
          & \quad + Text-guided & \textbf{0.303 } & \textbf{0.482 } \bigstrut[b]\\
    \hline
    \multirow{2}[2]{*}{$\mathcal{F}$ 20\%} & Pure Seg. & 0.357  & 0.562  \bigstrut[t]\\
          & \quad + Text-guided & \textbf{0.367 } & \textbf{0.575 } \bigstrut[b]\\
    \hline
    \multirow{2}[2]{*}{$\mathcal{F}$ 30\%} & Pure Seg. & 0.415  & 0.641  \bigstrut[t]\\
          & \quad + Text-guided & \textbf{0.438 } & \textbf{0.650 } \bigstrut[b]\\
    \hline
    \multirow{2}[2]{*}{$\mathcal{F}$ 40\%} & Pure Seg. & 0.452  & 0.660  \bigstrut[t]\\
          & \quad + Text-guided & \textbf{0.463 } & \textbf{0.667 } \bigstrut[b]\\
    \hline
    \end{tabular}%
    }}
  \end{minipage}%
  \hfill
  \begin{minipage}{0.40\linewidth}
  \vspace{0.3cm}
    \caption{Performance of step-1 tumor segmentation teacher network under various ratios of fully-supervised data ($\mathcal{F}$). We report Dice of tumors in the Esophagus (ESO) and Esophagogastric Junction (EGJ) on remaining weakly-supervised training images.}
    \label{tab:ratio1}%
  \end{minipage}
\end{table}

\begin{table}[!t]
\centering
  \scriptsize
  \caption{Performance of cancer detection with different proportions of fully-supervised data ($\mathcal{F}$) and weakly-supervised data ($\mathcal{W}$). We report AUC, sensitivity (Sens.), and specificity (Spec.) under several fully supervised settings ($\mathcal{F}$) and weakly semi-supervised settings ($\mathcal{F}$+$\mathcal{W}$).}
  \setlength{\tabcolsep}{1.9mm}{
    \begin{tabular}{c|l|ccc|ccc}
    \hline
    Label Types & Method & \multicolumn{3}{c|}{Val} & \multicolumn{3}{c}{Test} \bigstrut[t]\\
          &       & AUC   & Sens. & Spec. & AUC   & Sens. & Spec. \bigstrut[b]\\
    \hline
    \hline
    $\mathcal{F}$ 10\% & PANDA-SwinUNetrV2 & 0.857  & 0.625  & 0.946  & 0.841  & 0.643  & 0.867  \bigstrut[t]\\
    $\mathcal{F}$ 20\% & PANDA-SwinUNetrV2 & 0.902  & 0.750  & 0.893  & 0.874  & 0.719  & 0.859  \\
    $\mathcal{F}$ 30\% & PANDA-SwinUNetrV2 & 0.945  & 0.743  & 0.964  & 0.923  & 0.760  & 0.904  \\
    $\mathcal{F}$ 40\% & PANDA-SwinUNetrV2 & 0.961  & 0.790  & 0.938  & 0.952  & 0.827  & 0.933  \\
    $\mathcal{F}$ 100\% & PANDA-SwinUNetrV2 & \textbf{0.973 } & 0.934  & 0.929  & \textbf{0.966 } & 0.939  & 0.874  \bigstrut[b]\\
    \hline
    \hline
    \multirow{2}[2]{*}{$\mathcal{F}$ 10\% + $\mathcal{W}$ 90\%} & PANDA-SwinUNetrV2 & 0.938  & 0.763  & 0.938  & 0.934  & 0.776  & 0.948  \bigstrut[t]\\
          & Ours-SwinUNetrV2 & \textbf{0.951 } & 0.907  & 0.875  & \textbf{0.952 } & 0.908  & 0.867  \bigstrut[b]\\
    \hline
    \multirow{2}[2]{*}{$\mathcal{F}$ 20\% + $\mathcal{W}$ 80\%} & PANDA-SwinUNetrV2 & 0.930  & 0.868  & 0.871  & 0.944  & 0.867  & 0.882  \bigstrut[t]\\
          & Ours-SwinUNetrV2 & \textbf{0.958 } & 0.901  & 0.911  & \textbf{0.955 } & 0.934  & 0.859  \bigstrut[b]\\
    \hline
    \multirow{2}[2]{*}{$\mathcal{F}$ 30\% + $\mathcal{W}$ 70\%} & PANDA-SwinUNetrV2 & 0.953  & 0.895  & 0.938  & 0.946  & 0.867  & 0.896  \bigstrut[t]\\
          & Ours-SwinUNetrV2 & \textbf{0.966 } & 0.888  & 0.982  & \textbf{0.961 } & 0.903  & 0.904  \bigstrut[b]\\
    \hline
    \multirow{2}[2]{*}{$\mathcal{F}$ 40\% + $\mathcal{W}$ 60\%} & PANDA-SwinUNetrV2 & 0.962  & 0.855  & 0.973  & 0.955  & 0.852  & 0.926  \bigstrut[t]\\
          & Ours-SwinUNetrV2 & \textbf{0.974 } & 0.882  & 0.982  & \textbf{0.963 } & 0.878  & 0.948  \bigstrut[b]\\
    \hline
    \end{tabular}%
    }
  \label{tab:ratio2}%
\end{table}%

\begin{figure}[!t]
	\centering
 \begin{subfigure}{0.45\linewidth}
		\includegraphics[width=1\linewidth]{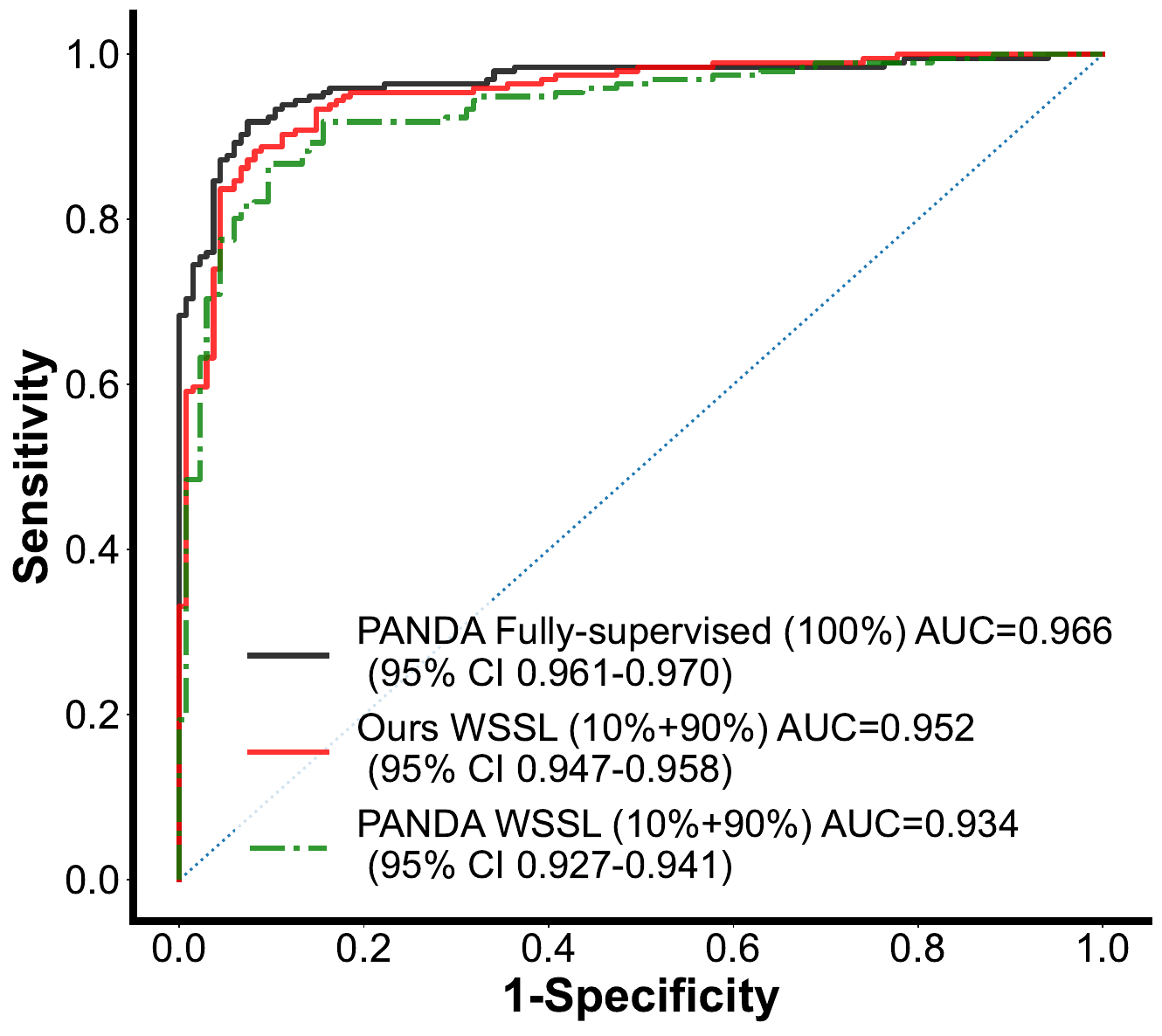}
		\label{fig:roc10}
		\caption{$\mathcal{F}$10\% + $\mathcal{W}$90\%}
	\end{subfigure}
 \hfill
	\begin{subfigure}{0.45\linewidth}
		\includegraphics[width=1\linewidth]{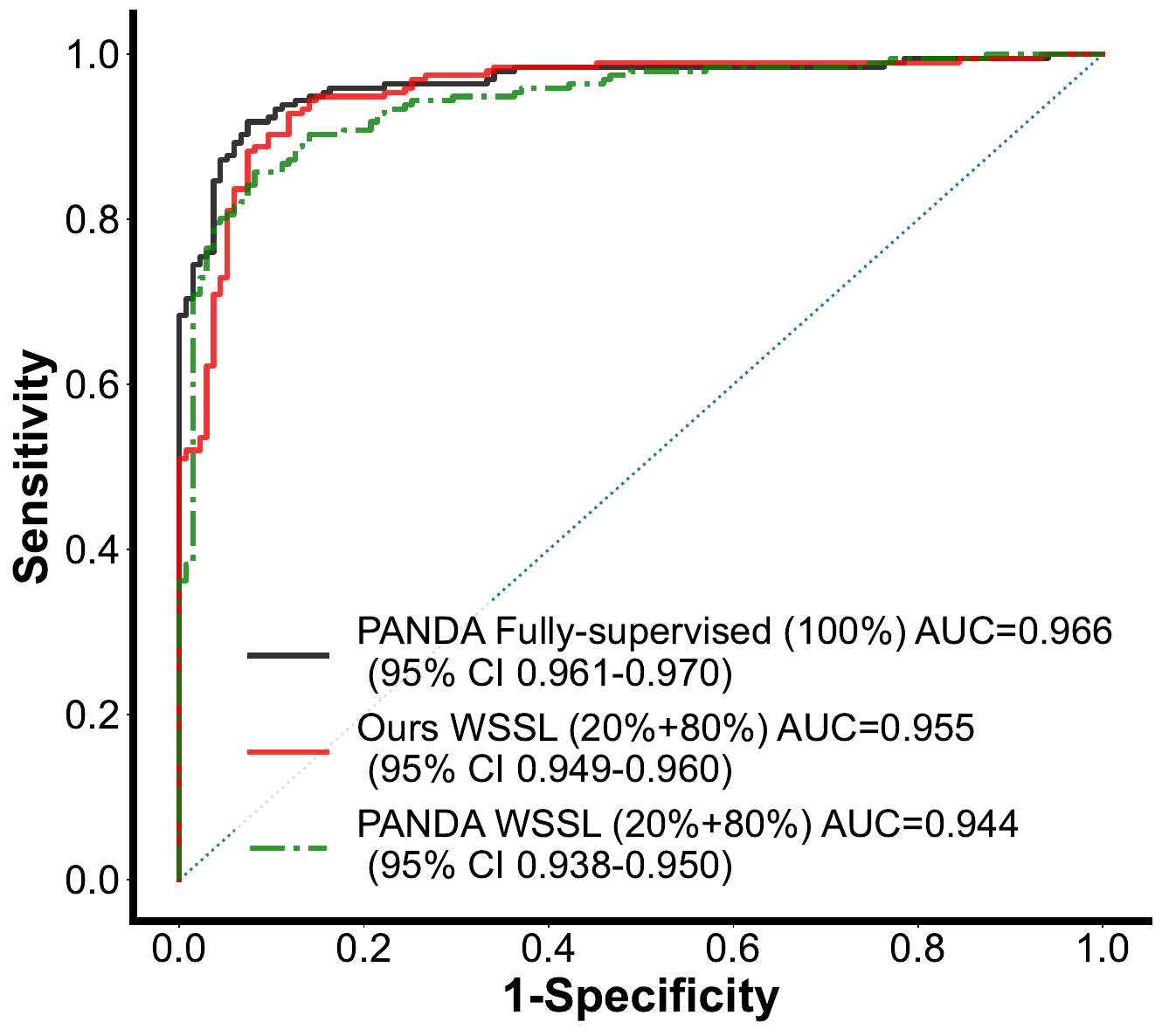}
		\label{fig:roc20}
		\caption{$\mathcal{F}$20\% + $\mathcal{W}$80\%}
	\end{subfigure}
	\begin{subfigure}{0.45\linewidth}
		\includegraphics[width=1\linewidth]{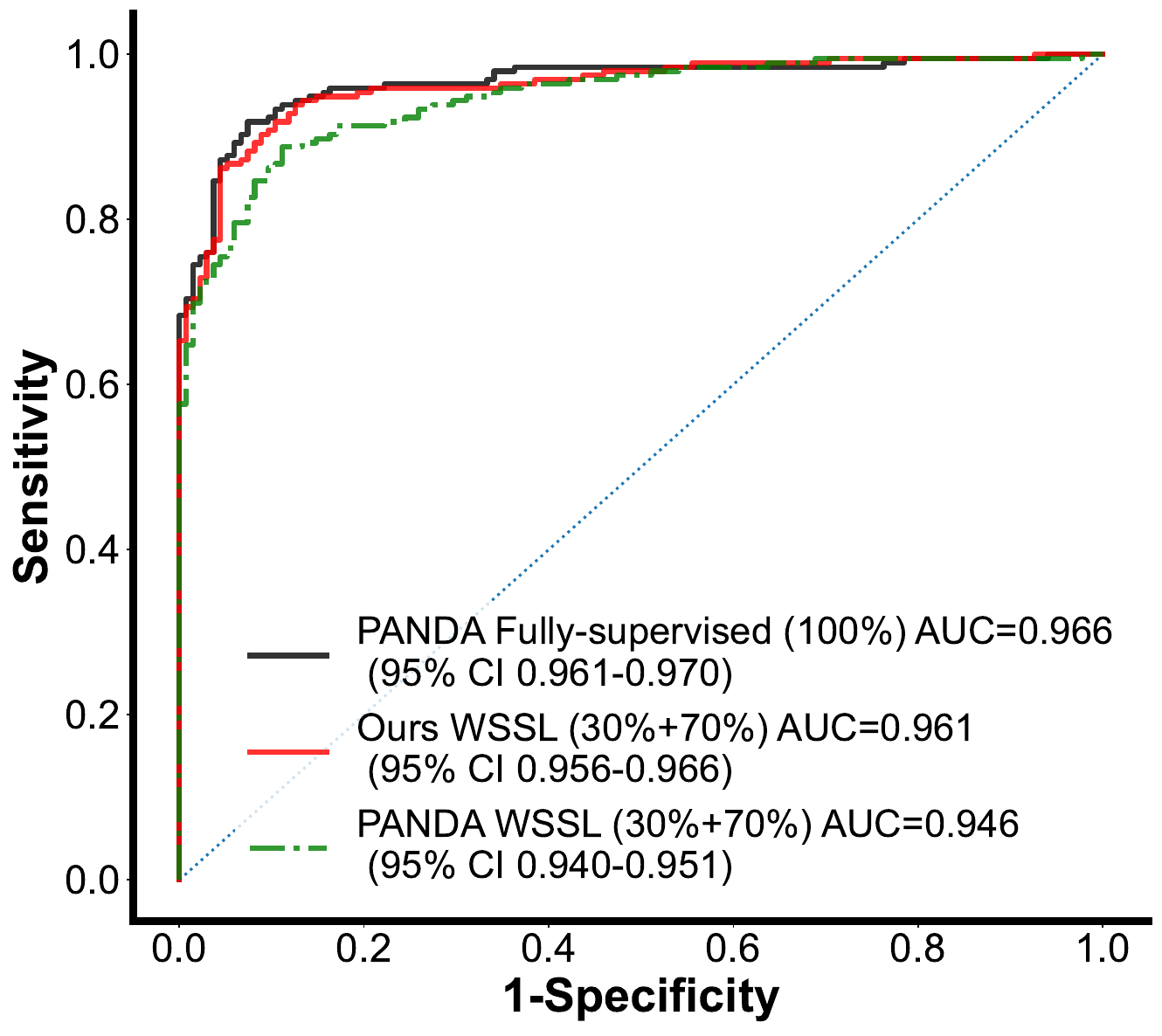}
		\label{fig:roc30}
		\caption{$\mathcal{F}$30\% + $\mathcal{W}$70\%}
	\end{subfigure}
 \hfill
	\begin{subfigure}{0.45\linewidth}
		\includegraphics[width=1\linewidth]{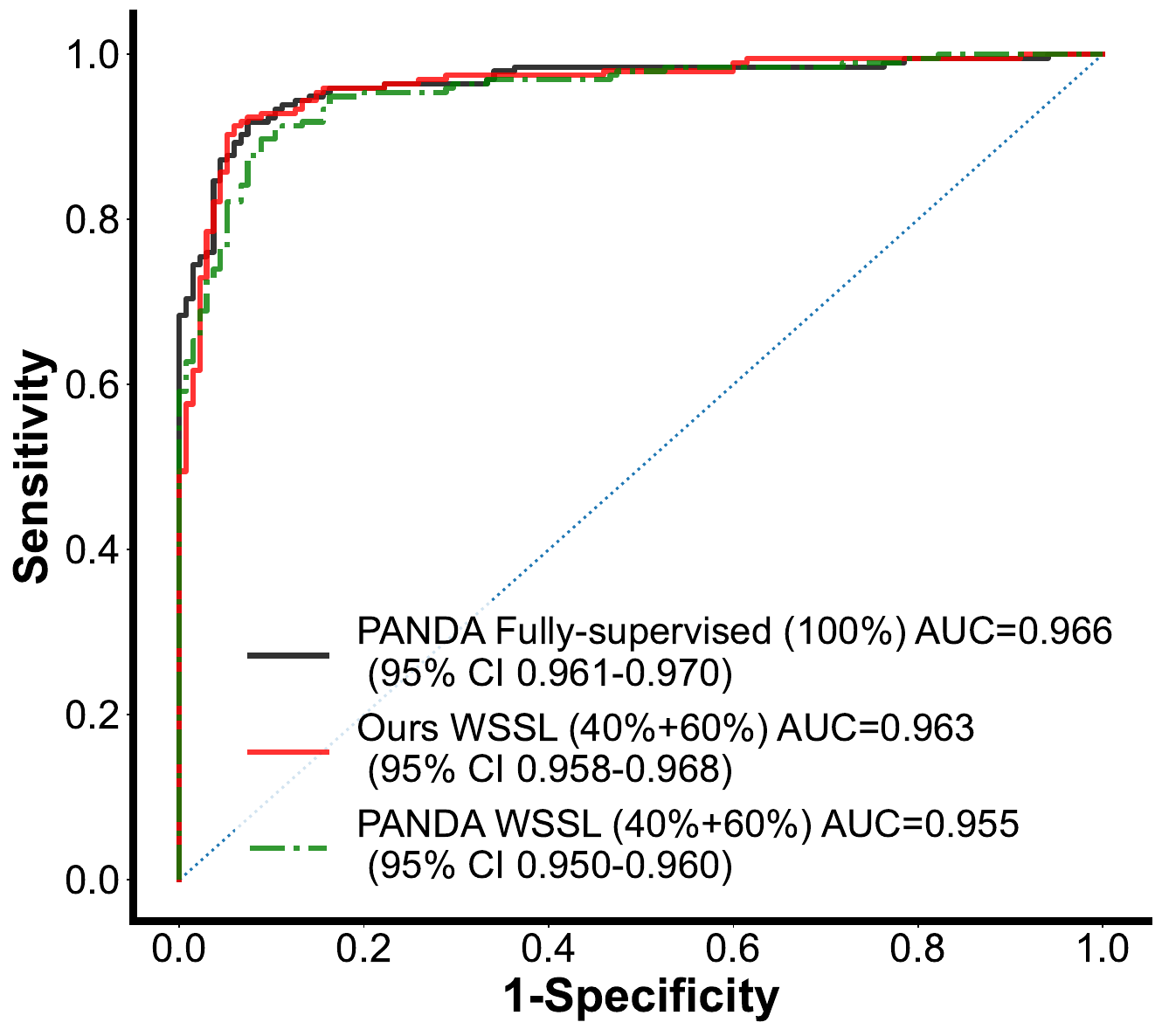}
		\label{fig:roc40}
		\caption{$\mathcal{F}$40\% + $\mathcal{W}$60\%}
	\end{subfigure}
	\caption{ROC curves of cancer detection under varying proportions of fully-supervised data ($\mathcal{F}$) and weakly-supervised data ($\mathcal{W}$).}
	\label{fig:roc_ratio}
\end{figure}

\noindent \textbf{A2. Step 2 cancer detection performance under different data proportions.}
In Table \ref{tab:ratio2}, we evaluate the effectiveness of our cancer detection method under different training data settings, under both fully-supervised scenarios and weakly semi-supervised scenarios.

In the pure fully-supervised setting ($\mathcal{F}$), SwinUNetrV2-based PANDA~\cite{cao2023large} is trained on labeled data, and we observe a direct link between the proportion of fully-supervised data and the performance metrics. Starting from $\mathcal{F}$ 20\% and advancing to $\mathcal{F}$ 100\%, there is a steady and significant increase in AUC, both on the validation and test sets, peaking at an impressive 0.973 and 0.966 respectively for the latter. Sensitivity and specificity also followed an upward trend, with the best figures attained at full supervision, highlighting the value of extensively labeled data in achieving optimal detection accuracy.

However, under the weakly semi-supervised setting ($\mathcal{F}$+$\mathcal{W}$) where the training involved a combination of both fully-supervised ($\mathcal{F}$) and weakly-supervised ($\mathcal{W}$) data, the model shows different degrees of improvement. Notably, integrating 80\% weakly-supervised data with 20\% fully-supervised data resulted in substantial gains in AUC for both the PANDA-SwinUNetrV2 and our enhanced version, labeled as `Ours-SwinUNetrV2'. Our approach outperforms the baseline with an AUC of 0.958 on the validation set and 0.955 on the test set, suggesting that our method effectively leverages the additional weakly-supervised data to improve performance.
As the ratio of fully-supervised data increased, our method continued to demonstrate superior results. With 40\% of the data fully labeled, our model reached an AUC of 0.974 on the validation set and 0.963 on the test set, again surpassing the baseline while maintaining high sensitivity and specificity. Notably, on the test set, we observe that when using 30\% of fully supervised data, the system can achieve commendable diagnostic performance with the fully supervised baseline model (0.961 vs. 0.966), and the DeLong test result $p=0.39$ indicates that our method using only 30\% tumor annotations is statistically equivalent to the PANDA model in 100\% fully-supervised setting. Finally, in Fig.~\ref{fig:roc_ratio}, we provide ROC curves of the proposed method and PANDA in different settings.

\section{Moreover comparision.}
\label{sec:comparision}

\noindent \textbf{B1. Text encoder initialization.} 
In this paper, since our model takes 3D images as input, we do not utilize the image encoder of VLM model, and only the text encoder is used. In the main paper, we utilize the text encoder from the most widely used VLM model CLIP~\cite{radford2021learning}.
In Table~ \ref{tab:medclip1} and \ref{tab:medclip2}, under 30\% fully supervised setting, we further compare with the text encoder of MedCLIP~\cite{wang2022medclip}, which achieves similarity performance to that of the CLIP, showing AUC scores of 0.959 and 0.961 on the test set, respectively. The Delong test yielded a p-value of 0.801, indicating no significant difference between the two models. Considering CLIP is also trained on many medical image-text pairs, its text encoder is already well-attuned to understand the context of medical words. 

\begin{table}[!t]
  \centering
  \begin{minipage}{0.4\linewidth} 
    \centering
    \scriptsize
    \caption{Dice of tumor segmentation student on the remain 70\% weakly-supervised training set.}
    \label{tab:medclip1}%
    \setlength{\tabcolsep}{0.2mm}{
      \renewcommand{\arraystretch}{0.6}{
    \begin{tabular}{l|cc}
    \hline
    Method & ESO   & EGJ \bigstrut\\
    \hline
    Pure Seg. & 0.415  & 0.641  \bigstrut[t]\\
    \quad + Text-guided (MedCLIP) & 0.436  & 0.654  \\
    \quad + Text-guided (CLIP) & 0.438  & 0.650  \bigstrut[b]\\
    \hline
    \end{tabular}%
    }}
  \end{minipage}%
  \hfill
  \begin{minipage}{0.55\linewidth}
    \centering
    \scriptsize
    \caption{Performance of step-2 cancer detection student on the validation and test set under 30\%$\mathcal{F}$ +70\%$\mathcal{W}$ setting.}
    \label{tab:medclip2}%
    \setlength{\tabcolsep}{0.2mm}{
      \renewcommand{\arraystretch}{0.6}{
    \begin{tabular}{l|ccc|ccc}
    \hline
    Method & \multicolumn{3}{c|}{Val} & \multicolumn{3}{c}{Test} \bigstrut[t]\\
          & AUC   & Sens. & Spec. & AUC   & Sens. & Spec. \bigstrut[b]\\
    \hline
    Ours (MedCLIP) & \textbf{0.966 } & 0.888  & 0.938  & 0.959  & 0.918  & 0.859  \bigstrut[t]\\
    Ours (CLIP) & \textbf{0.966 } & 0.888  & 0.982  & \textbf{0.961 } & 0.903  & 0.904  \bigstrut[b]\\
    \hline
    \end{tabular}%
    }}
  \end{minipage}
\end{table}

\noindent \textbf{B1. Compare with Bosma \etal~\cite{bosma2023semisupervised}.} 
Bosma \etal~\cite{bosma2023semisupervised} present a report-guided semi-supervised model for the prostate cancer, which designs a post-processing strategy for pseudo masks informed according to lesion counts from prostate cancer reports. We draw inspiration from Bosma \etal~\cite{bosma2023semisupervised} to select pseudo masks that match locations from reports. As shown in Table~\ref{tab:bosma}, ``PANDA-SwinUNetrV2+\cite{bosma2023semisupervised}'' results in marginal improvements from PANDA, which demonstrates our method can better refine the semi-supervised paradigm by incorporating infroamtion from clinical reports via a learning-based method that undertakes cross-modality learning. This learnable process in both tumor segmentation teacher and cancer detection student enables the model to correct inaccuracies in pseudo tumor masks, ensuring their alignment with the tumor locations as detailed in clinical reports.

\begin{table}[!t]
\centering
  \scriptsize
  \caption{Comparision with Bosma \etal~\cite{bosma2023semisupervised} under 30\% .}
  \setlength{\tabcolsep}{1.5mm}{
    \begin{tabular}{c|l|ccc|ccc}
    \hline
    Label Types & Method & \multicolumn{3}{c|}{Val} & \multicolumn{3}{c}{Test} \bigstrut[t]\\
          &       & AUC   & Sens. & Spec. & AUC   & Sens. & Spec. \bigstrut[b]\\
    \hline
    \hline
          & PANDA-SwinUNetrV2 & 0.953  & 0.895  & 0.938  & 0.946  & 0.867  & 0.896  \bigstrut[t]\\
    30\%$\mathcal{F}$ +70\%$\mathcal{W}$ & PANDA-SwinUNetrV2+\cite{bosma2023semisupervised} & 0.954  & 0.842  & 0.964  & 0.948  & 0.947  & 0.830  \\
          & Ours-SwinUNetrV2 & \textbf{0.966 } & 0.888  & 0.982  & \textbf{0.961 } & 0.903  & 0.904  \bigstrut[b]\\
    \hline
    \end{tabular}%
    }
  \label{tab:bosma}%
\end{table}%

\end{document}